\documentclass[letterpaper, 10 pt, conference]{ieeeconf}
\IEEEoverridecommandlockouts  
\usepackage{times,multirow,caption,float}
\usepackage{wrapfig}
\usepackage{graphicx}
\usepackage{microtype}
\usepackage{svg}
\usepackage{epsfig,xspace,layout}
\usepackage[ruled]{algorithm2e}
\usepackage{color}
\usepackage{amsfonts}
\usepackage{times}
\usepackage{amssymb}
\usepackage{amsmath}
\usepackage{rotating}

\usepackage{amsthm}
\usepackage{graphicx}
\usepackage{epsfig}
\usepackage{mathrsfs}
\usepackage{mathtools}
\usepackage{makeidx}
\usepackage{multirow} 
\usepackage{threeparttable}
\usepackage{dsfont}
\usepackage{tikz}
\usetikzlibrary{shapes,arrows,positioning,calc}
\usepackage{siunitx}
\usepackage{url}
\usepackage{cite}

\usepackage{hyperref}
\hypersetup{
    colorlinks=true,
    linkcolor=blue,
    filecolor=blue,      
    urlcolor=blue,
    citecolor=cyan,
}
\usepackage[noabbrev]{cleveref}
\usepackage[font={small}]{caption}
\usetikzlibrary{arrows.meta}
\usetikzlibrary{calc}
\usepackage{booktabs}
\usepackage{pgfplots}
\pgfplotsset{compat=1.3}

\theoremstyle{definition}

\theoremstyle{remark}

\DeclareMathAlphabet{\mathpzc}{OT1}{pzc}{m}{it}

\DeclareFontFamily{U}{jkpmia}{}
\DeclareFontShape{U}{jkpmia}{m}{it}{<->s*jkpmia}{}
\DeclareFontShape{U}{jkpmia}{bx}{it}{<->s*jkpbmia}{}
\DeclareMathAlphabet{\mathfrak}{U}{jkpmia}{m}{it}

\newcommand{\R}{\mathbb{R}}

\newcommand\figprocess{
\begin{figure}[t!]
    \centering
    \resizebox{1.\linewidth}{!}{
    \begin{tikzpicture}
    \draw[fill=black] (-1.8,0) rectangle(2.2,-0.01);
    \node[scale=0.5] at (-1.4,-0.15) {$t_j$};
    \node[scale=0.5] at (-1.0,-0.15) {$t_t$};
    \node[scale=0.5] at (0.0,-0.15) {$t_h$};

    \draw[my_blue!40, dashed]  (-1,0.15) parabola bend (0,0.7) (0.7,0.15) ;
    \draw[my_blue, very thick] (-1.4,0.05) -- +(0:0.3) -- +(0:-0.3) node[black, above, scale=0.5] {Jump start};
    \draw[my_blue] (-1.6,0.05) -- +(-170:0.2);
    \draw[my_blue] (-1.2,0.05) -- +(-170:0.2);

    \draw[my_blue!50, very thick] (-1,0.15) -- +(10:0.3) -- +(10:-0.3);
    \draw[black, -{Stealth[scale=0.5]}](-1,0.15) -- (-0.8,0.4) node[left, scale=0.5] {Takeoff};
    \draw[my_blue!50] ([shift=(10:0.25)] -1,0.15) -- +(-150:0.2)  -- +(0,-0.2);
    \draw[my_blue!50] ([shift=(10:-0.25)] -1,0.15) -- +(-150:0.2)  -- +(0,-0.1);
    \draw[-{Stealth[scale=0.4]}] (-1.1,0.15) arc (-180:0:0.1);
    \draw[my_blue, very thick] (0,0.7) -- +(-15:0.3) -- +(-15:-0.3) node[black, above, scale=0.5] {Collision};
    \draw[my_blue] ([shift=(-15:0.25)] 0,0.7) -- +(-150:0.2)  -- +([shift=(-150:0.2)] -40:0.2);
    \draw[my_blue] ([shift=(-15:-0.25)] 0,0.7) -- +(-150:0.2)  -- +([shift=(-150:0.2)] -40:0.2);
    \draw[my_blue!40, very thick] (0.7,0.18) -- +(-25:0.3) -- +(-25:-0.3) ;
    \draw[my_blue!40] ([shift=(-25:0.25)] 0.7,0.18) -- +(-170:0.2)  -- +(0,-0.08);
    \draw[my_blue!40] ([shift=(-25:-0.25)] 0.7,0.18) -- +(-120:0.18)  -- +(0,-0.28);

    \draw[my_red] ([shift=(-15:0.18)] 0,0.7) +(75:0.05) circle (0.7pt);
    \draw[my_red] (1.5,0.05) circle (0.7pt);
    \draw[my_red, dashed]  ([shift=(-15:0.18)] 0,0.7) +(75:0.05) parabola bend (0.5,0.8) (1.5,0.05) node[black, above right, scale=0.5] {Landing point $t_e$};
    \draw[my_red] (2,0.8) circle (0.7pt) node[black,right,scale=0.5,align=left] {Initial position \\ and velocity};
    \draw[my_red, dashed]  ([shift=(-15:0.18)] 0,0.7) +(75:0.05) parabola bend (1.2,1.2) (2,0.8) ;
    \draw[black,-{Stealth[scale=0.5]}](2,0.8) -- (1.8,1.) ;
    \end{tikzpicture}
    }
    \caption{The process of ball bumping with the quadruped robot.}
    \label{fig:bbq_process}
\end{figure}
}

\newcommand\figdemo{
\begin{figure}[t!]
    \centering
    \includegraphics[width=0.4\linewidth]{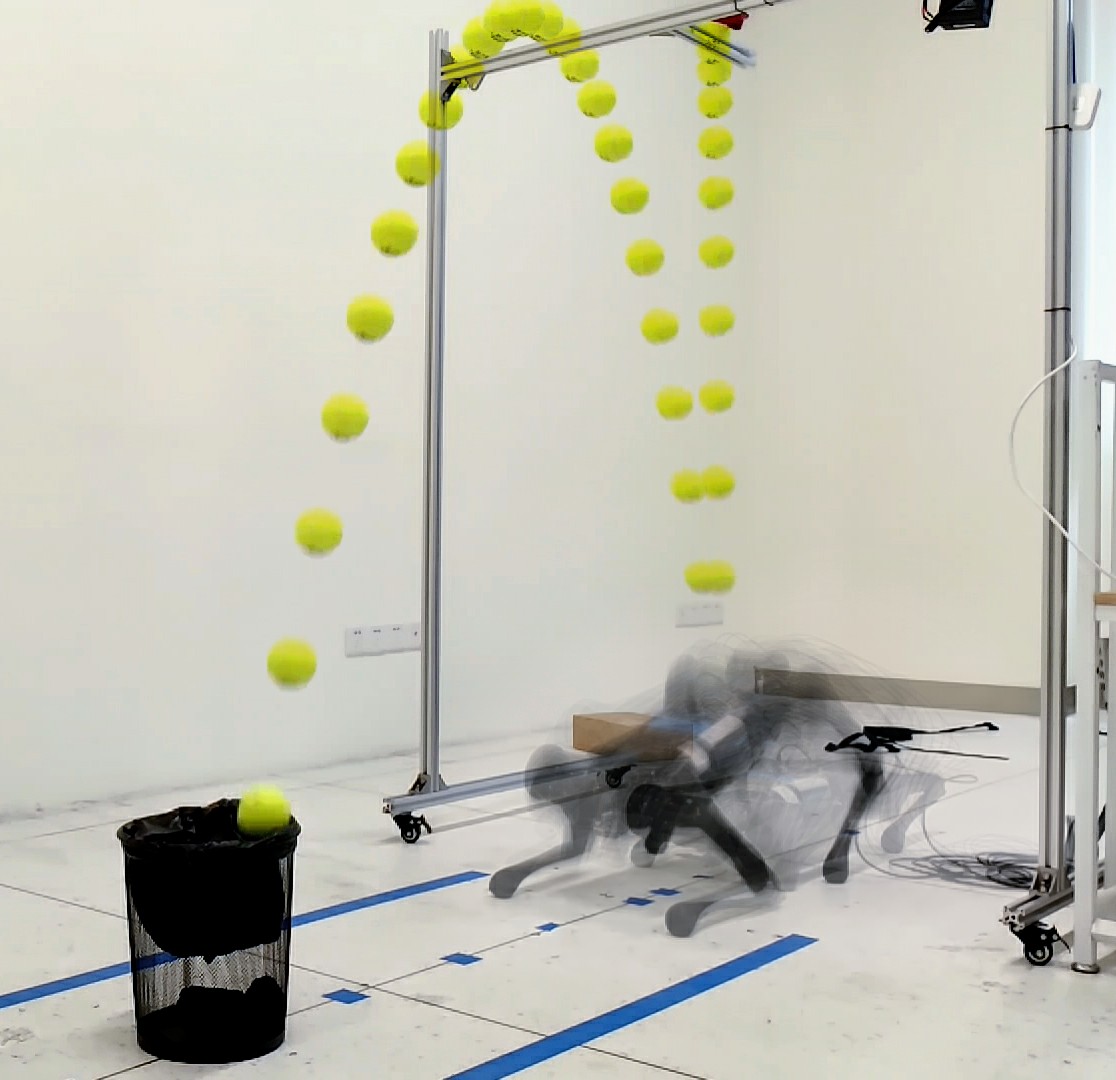}
    \figsetup
    \caption{\textbf{Left:} A quadruped robot bumps the ball into a trash can. \textbf{Right:} experiment setup.}
    \label{fig:exp_demo_setup}
    \vspace{-15pt}
\end{figure}
}

\newcommand\figplanarmodel{
    \begin{tikzpicture}
    \draw[fill=black] (-0.3,0) rectangle(0.3,-0.01);
    \filldraw[draw=my_blue,line width=0.05mm, fill=my_blue!20, shift ={(-0.3, 0.2),(0 ,0.3)}, rotate around={-30:(0.3,0.1)}] (0,0) rectangle (0.6,0.15);
    \draw[my_blue, line width=0.05mm] (0.25,0) -- (0.1,0.05) -- (0.2,0.07);
    \draw[my_blue, line width=0.05mm] (-0.2,0) -- (-0.3,0.2) -- (-0.29,0.36);
    \draw[my_red, -{Stealth[scale=0.2]}, line width=0.1mm] (-0.2,0) -- (-0.11,0.2) node[right,scale=0.2]{$\mathbf{f}_r$} ;
    \draw[my_red, -{Stealth[scale=0.2]}, line width=0.1mm] (0.25,0) -- (0.3,0.15) node[right,scale=0.2]{$\mathbf{f}_f$} ;
    \draw[my_orange, dash pattern=on 1pt off 0.5pt, line width=0.05mm] (0,0.28) -- (0.2,0.28) ;
    \draw[my_orange, line width=0.05mm] (0,0.28) -- +(-30:0.2) node[xshift=-1pt,yshift=2pt,scale=0.2]{$\theta$};
    \draw[my_orange, line width=0.05mm] ([shift={(0,0.28)}] 0.1,0) arc (0:-30:0.1);
    \end{tikzpicture}
}

\newcommand\fighitvel{
    \begin{tikzpicture}
    \filldraw[draw=my_blue, fill=my_blue!20, rounded corners, shift ={(-5.0 ,-0.5),(0 ,3)}, rotate around={-30:(3,0.75)}] (0,0) rectangle (6,1.5);
    \draw[my_red, -Stealth, line width=0.5mm] (80:3) node[left,scale=1.5]{$\dot{\mathbf{x}}_b(t_{h_-})$} -- (0,0) ;
    \draw[my_red, -Stealth, line width=0.5mm] (0,0) -- ([shift=(35:2.25)] 60:1.5) node[right,scale=1.5]{$\dot{\mathbf{x}}_b(t_{h_+})$};
    \draw[-{Stealth[scale=1]}] ([shift={(-2,0.3)}] -30:-0.5) arc (-210:-30:0.5);
    \draw[-{Stealth[scale=1]}] (-2,0.3) node[above,scale=1.5]{$\dot{\mathbf{x}}_q(t_h)$} --+(20:1) ;
    \draw[my_orange, line width=0.5mm] ([shift={(-2,0.3)}] 0,0) -- (-120:0.75) node[left, scale=1.5]{$l_h$};
    \draw[my_orange, line width=0.5mm] (0,0) -- (60: -0.75) node[right, scale=1.5]{$h_{body}$};
    \draw[-Stealth, line width=0.5mm] (0,0) -- (50:2) node[xshift=3mm, yshift=2mm, scale=1.5]{$\mathbf{v}_p$};
    
    \draw[dotted, line width=0.5mm] (0,0) -- (60:3);
    \end{tikzpicture}
}

\newcommand\figsetup{
    \resizebox{0.55\linewidth}{!}{
    \begin{tikzpicture}
        \node at (0,0) {\includegraphics[width=0.8\linewidth]{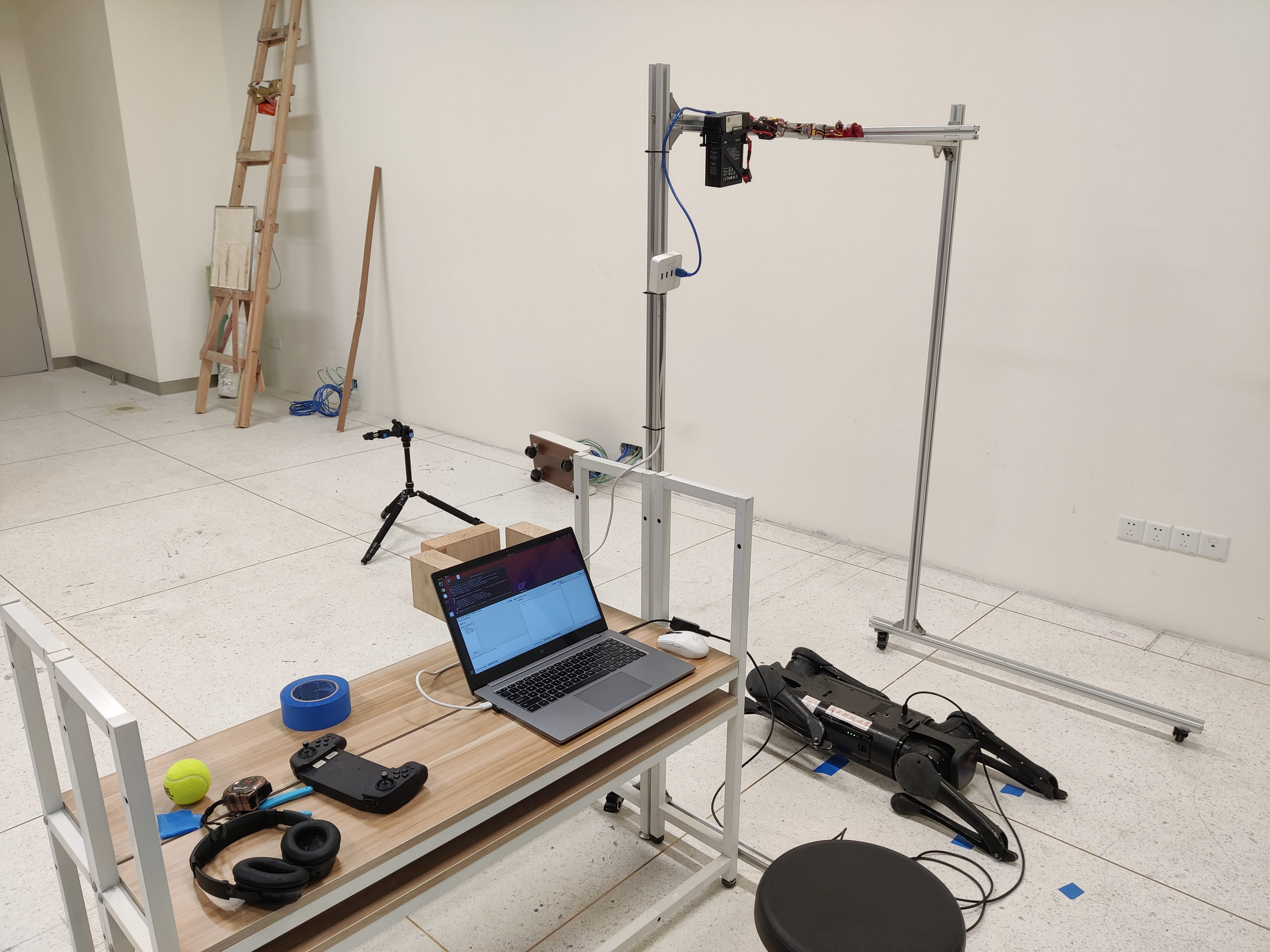}};
        \node at(-2.5,2) [line width=0.2mm, draw=my_blue, inner sep=0.2] {\includegraphics[width=0.5\linewidth]{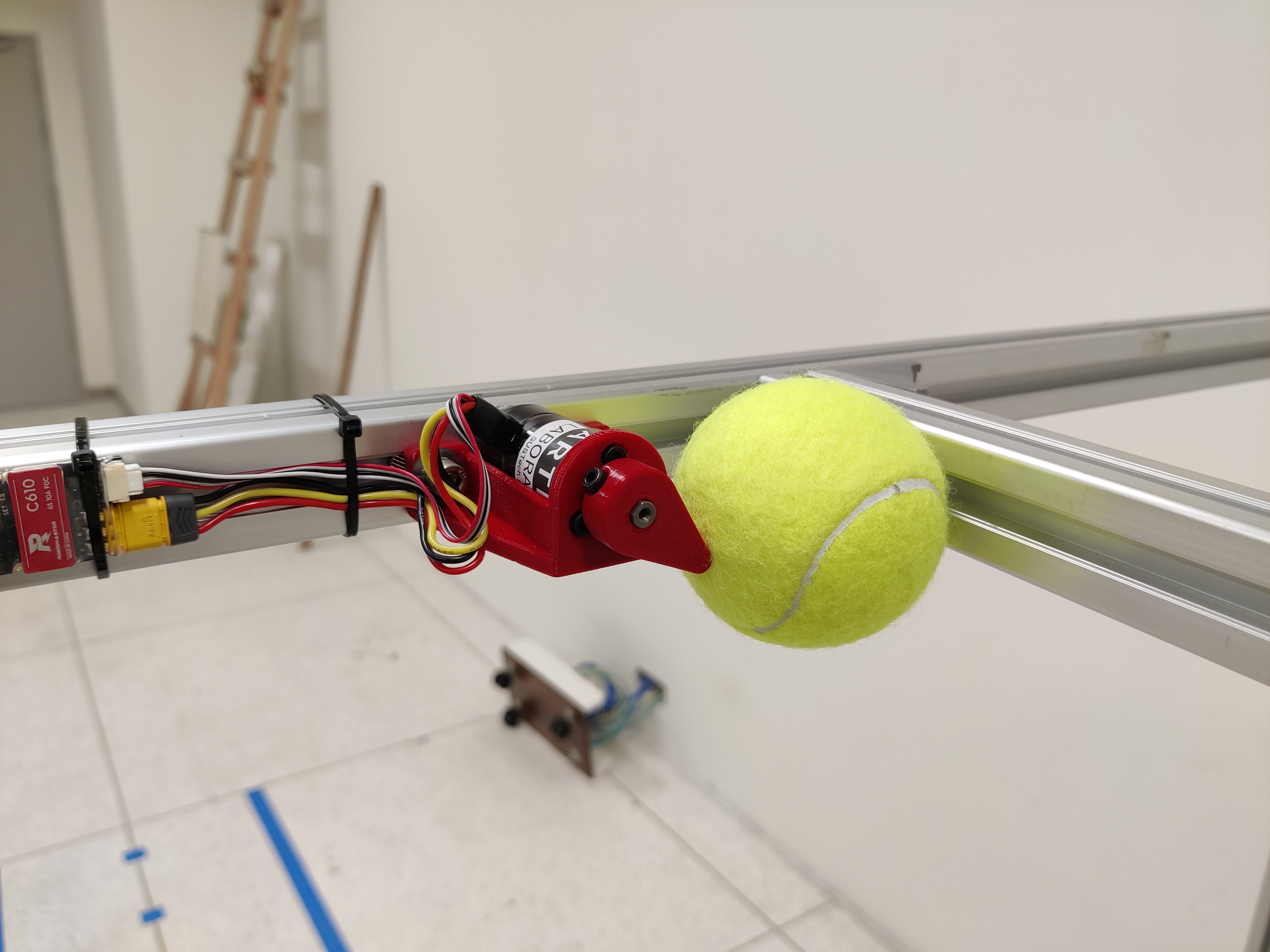}};
        
        \draw[line width=0.2mm, draw=my_blue] (0.9,1.6) rectangle (1.4,2.1);
        \draw[line width=0.2mm, draw=my_blue] (0.9,1.6) -- (-0.34, 0.36);
        \draw[line width=0.2mm, draw=my_blue] (0.9,2.1) -- (-0.34, 3.62);
    
        \draw[latex-, very thick, draw=my_blue] (-1,-0.7) -- ++(-0.5,0) node[left, black]{\small Laptop};
        \draw[latex-, very thick, draw=my_blue] (1.8,-1.2) -- ++(0.5,0.5) node[above, black]{\small Aliengo};
        
        \draw[latex-, very thick, draw=my_blue] (-2.0,2.5) -- ++(0,0.5) node[above, black]{\small Tennis Ball};
        \draw[latex-, very thick, draw=my_blue] (-3.2,1.8) -- ++(210:0.8) node[below, black]{\small Releaser};
    \end{tikzpicture}
    }
}

\definecolor{my_blue}{rgb}{0, 0.4470, 0.7410}
\definecolor{my_orange}{rgb}{0.8500, 0.3250, 0.0980}
\definecolor{my_yellow}{rgb}{0.9290, 0.6940, 0.1250}
\definecolor{my_purple}{rgb}{0.4940, 0.1840, 0.5560}
\definecolor{my_green}{rgb}{0.4660, 0.6740, 0.1880}
\definecolor{my_red}{rgb}{0.6350, 0.0780, 0.1840}
\definecolor{my_black}{rgb}{0.25, 0.25, 0.25}

\newcommand\plottraj{
\begin{figure}[t!]
    \centering
    \begin{tikzpicture}
    \pgfplotsset{
        xmin=0.196, xmax=0.78,
        ymin=-0.05, ymax=0.3,
        width=0.85\linewidth, height=0.6\linewidth, ymajorgrids=true, grid style=dashed, legend pos=north west, legend columns=3, 
        label style={font=\scriptsize}, tick label style={font=\scriptsize}
    }
    \begin{axis}[axis y line*=left, xlabel = {Time [s]}, ylabel = {Position [m]}]
        \addplot[my_blue] table[col sep=comma, x = time_croped, y = /odom/pose/pose/position/x] {data/trash_odom.csv}; \addlegendentry{$\mathbf{x}_{q,x}$};
        \addplot[my_blue,dashed] table[col sep=comma, x = time, y = pose/x] {data/traj.csv}; \addlegendentry{$\mathbf{x}^{des}_{q,x}$};
        \addplot[my_red] table[col sep=comma, x = time_croped, y = /odom/pose/pose/position/z] {data/trash_odom.csv}; \addlegendentry{$\mathbf{x}_{q,z}$};
        \addplot[my_red,dashed] table[col sep=comma, x = time, y = pose/z] {data/traj.csv}; \addlegendentry{$\mathbf{x}^{des}_{q,z}$};
        
        \addlegendimage{/pgfplots/refstyle=plot:pitch}\addlegendentry{$\mathbf{x}_{q,\theta}$};
        \addlegendimage{/pgfplots/refstyle=plot:pitch_des}\addlegendentry{$\mathbf{x}^{des}_{q,\theta}$};
    \end{axis}
    \begin{axis}[axis y line*=right, axis x line=none, ylabel = {Pitch angle [rad]}]
        \addplot[my_yellow] table[col sep=comma, x = time_croped, y = pitch] {data/trash_odom.csv}; \label{plot:pitch}
        \addplot[my_yellow,dashed] table[col sep=comma, x = time, y = pose/theta] {data/traj.csv}; \label{plot:pitch_des}
        
        \addplot[dashed] coordinates{(0.691,-0.1)(0.691,0.3)};
        \addplot[dashed] coordinates{(0.262,-0.1)(0.262,0.2)};
        \addplot[dashed] coordinates{(0.322,-0.1)(0.322,0.2)};
    \end{axis}
    \end{tikzpicture}
    
    \begin{tikzpicture}
        \pgfplotsset{
        xmin=0.196, xmax=0.78,
        ymin=-0.1, ymax=1.6,
        width=0.85\linewidth, height=0.6\linewidth, ymajorgrids=true, grid style=dashed, legend pos=north west, legend columns=3, 
        label style={font=\scriptsize}, tick label style={font=\scriptsize}
    }
    \begin{axis}[axis y line*=left, xlabel = {Time [s]}, ylabel = {Linear Velocity [m/s]}]

        \addplot[my_blue] table[col sep=comma, x = time_croped, y = /odom/twist/twist/linear/x] {data/trash_odom.csv}; \addlegendentry{$\dot{\mathbf{x}}_{q,x}$};
        \addplot[my_blue,dashed] table[col sep=comma, x = time, y = twist/x] {data/traj.csv}; \addlegendentry{$\dot{\mathbf{x}}^{des}_{q,x}$};
        \addplot[my_red] table[col sep=comma, x = time_croped, y = /odom/twist/twist/linear/z] {data/trash_odom.csv}; \addlegendentry{$\dot{\mathbf{x}}_{q,z}$};   
        \addplot[my_red,dashed] table[col sep=comma, x = time, y = twist/z] {data/traj.csv}; \addlegendentry{$\dot{\mathbf{x}}^{des}_{q,z}$};

        \addlegendimage{/pgfplots/refstyle=plot:pitch_dot}\addlegendentry{$\dot{\mathbf{x}}_{q,\theta}$};
        \addlegendimage{/pgfplots/refstyle=plot:pitch_dot_des}\addlegendentry{$\dot{\mathbf{x}}^{des}_{q,\theta}$};
    \end{axis}
    \begin{axis}[axis y line*=right, axis x line=none, ylabel = {Angular Velocity [rad/s]}]
        \addplot[my_yellow] table[col sep=comma, x = time_croped, y = /odom/twist/twist/angular/y] {data/trash_odom.csv}; \label{plot:pitch_dot}
        \addplot[my_yellow,dashed] table[col sep=comma, x = time, y = twist/omega] {data/traj.csv}; \label{plot:pitch_dot_des}
        
        \addplot[dashed] coordinates{(0.691,0)(0.691,1.6)};
        \addplot[dashed] coordinates{(0.262,-0.2)(0.262,0.2)};
        \addplot[dashed] coordinates{(0.322,-0.2)(0.322,0.2)};
    \end{axis}
    
    \end{tikzpicture}

    \caption{Planned trajectory and state estimate data in bump ball into a trash can experiment. The initial state of  The desired takeoff state is $\mathbf{x}_q^{des}(t_t) = [0.1\ 0.216\ 0.1\ 0.948\ 1.6\ 0.5]^T$, the actual takeoff state is $\mathbf{x}_q(t_t) = [0.088\ 0.235\ 0.087\ 0.86\ 1.5\ 0.516]^T$.}
    \label{fig:exp_traj}
\end{figure}
}

\newcommand\figforce{
\begin{figure}[t!]
    \centering
    \begin{tikzpicture}
    \begin{axis}[width=0.49\linewidth, height=0.4\linewidth, ymajorgrids=true, grid style=dashed, xmin=0.196, xmax=0.8, xlabel = {Time [s]}, ylabel = {GRF [N]},
    label style={font=\scriptsize}, tick label style={font=\scriptsize}, x label style={at={(axis description cs:0.5,-0.15)}}, y label style={at={(axis description cs:-0.3,.5)}}]
        \addplot[dashed] coordinates{(0.262,-10)(0.262,20)};
        \addplot[dashed] coordinates{(0.322,-10)(0.322,20)};
        \addplot[dashed] coordinates{(0.691,-10)(0.691,160)};
        \addplot[my_blue] table[col sep=comma, x = time_croped, y = /leg_states/reaction_force.0/x] {data/force.csv};
        \addplot[my_orange] table[col sep=comma, x = time_croped, y = /leg_states/reaction_force.0/y] {data/force.csv};        
        \addplot[my_yellow] table[col sep=comma, x = time_croped, y = /leg_states/reaction_force.0/z] {data/force.csv};
    \end{axis}
    \end{tikzpicture}
    \begin{tikzpicture}
    \begin{axis}[width=0.49\linewidth, height=0.4\linewidth, ymajorgrids=true, grid style=dashed, xmin=0.196, xmax=0.8, xlabel = {Time [s]}, ylabel = {GRF [N]},
    label style={font=\scriptsize}, tick label style={font=\scriptsize}, x label style={at={(axis description cs:0.5,-0.15)}}, y label style={at={(axis description cs:-0.3,.5)}}]
        \addplot[dashed] coordinates{(0.262,-10)(0.262,20)};
        \addplot[dashed] coordinates{(0.322,-10)(0.322,20)};
        \addplot[dashed] coordinates{(0.691,-10)(0.691,160)};
        \addplot[my_blue] table[col sep=comma, x = time_croped, y = /leg_states/reaction_force.1/x] {data/force.csv};
        \addplot[my_orange] table[col sep=comma, x = time_croped, y = /leg_states/reaction_force.1/y] {data/force.csv};        
        \addplot[my_yellow] table[col sep=comma, x = time_croped, y = /leg_states/reaction_force.1/z] {data/force.csv};
    \end{axis}
    \end{tikzpicture}
    \begin{tikzpicture}
    \begin{axis}[width=0.49\linewidth, height=0.4\linewidth, ymajorgrids=true, grid style=dashed, xmin=0.196, xmax=0.8, xlabel = {Time [s]}, ylabel = {GRF [N]},
    label style={font=\scriptsize}, tick label style={font=\scriptsize}, x label style={at={(axis description cs:0.5,-0.15)}}, y label style={at={(axis description cs:-0.3,.5)}}]
        \addplot[dashed] coordinates{(0.262,-10)(0.262,20)};
        \addplot[dashed] coordinates{(0.322,-10)(0.322,20)};
        \addplot[dashed] coordinates{(0.691,-10)(0.691,160)};
        \addplot[my_blue] table[col sep=comma, x = time_croped, y = /leg_states/reaction_force.2/x] {data/force.csv};
        \addplot[my_orange] table[col sep=comma, x = time_croped, y = /leg_states/reaction_force.2/y] {data/force.csv};        
        \addplot[my_yellow] table[col sep=comma, x = time_croped, y = /leg_states/reaction_force.2/z] {data/force.csv};
    \end{axis}
    \end{tikzpicture}
    \begin{tikzpicture}
    \begin{axis}[width=0.49\linewidth, height=0.4\linewidth, ymajorgrids=true, grid style=dashed, xmin=0.196, xmax=0.8, xlabel = {Time [s]}, ylabel = {GRF [N]},
    label style={font=\scriptsize}, tick label style={font=\scriptsize}, x label style={at={(axis description cs:0.5,-0.15)}}, y label style={at={(axis description cs:-0.3,.5)}}]
        \addplot[dashed] coordinates{(0.262,-10)(0.262,20)};
        \addplot[dashed] coordinates{(0.322,-10)(0.322,20)};
        \addplot[dashed] coordinates{(0.691,-10)(0.691,160)};
        \addplot[my_blue] table[col sep=comma, x = time_croped, y = /leg_states/reaction_force.3/x] {data/force.csv};
        \addplot[my_orange] table[col sep=comma, x = time_croped, y = /leg_states/reaction_force.3/y] {data/force.csv};        
        \addplot[my_yellow] table[col sep=comma, x = time_croped, y = /leg_states/reaction_force.3/z] {data/force.csv};
    \end{axis}
    \end{tikzpicture}
    \caption{The command ground reaction force (GRF) computed by jumping controller. the figures from left to right, top to bottom, represent FL, FR, RL, RR respectively. The values of $f_x$ $f_y$ $f_z$ axis are represented by blue, orange, and yellow respectively.}
    \label{fig:exp_force}
\end{figure}
}

\title{Real-time Trajectory Optimization and Control for Ball Bumping with Quadruped Robots}
\author{Qiayuan Liao, Zhefeng Cao, Hua Chen, and Wei Zhang
\thanks{All authors are with the Department of Mechanical and Energy Engineering, Southern University of Science and Technology, Shenzhen, China. {\tt liaoqiayuan@gmail.com, 12150041@mail.sustech.edu.cn, chenh6@sustech.edu.cn, zhangw3@sustech.edu.cn}. 
Qiayuan Liao is also with the School of Electromechanical Engineering, Guangdong University of Technology, Guangzhou, China.}
}

\begin{document}

\maketitle

\begin{abstract}    
This paper studies real-time motion planning and control for ball bumping motion with quadruped robots. To enable the quadruped to bump the flying ball with different initializations, we develop a nonlinear trajectory optimization-based planning scheme that jointly identifies the take-off time and state to achieve accurate ball hitting during the flight phase. Such a planning scheme employs a two-dimensional single rigid body model that achieves a satisfactory balance between accuracy and efficiency for the highly time-sensitive task. To precisely execute the planned motion, the tracking controller needs to incorporate the strict time-state constraint imposed on the take-off and ball-hitting events. To this end, we develop an improved model predictive controller that respects the critical time-state constraints. The proposed planning and control framework is validated with a real Aliengo robot. Experiments show that the problem planning approach can be computed in approximately $60$ ms on average, enabling successful accomplishment of the ball bumping motion with various initializations in real-time.

\end{abstract}
\section{Introduction}
Jumping and interacting with objects in the air is one of the most amazing behaviors that animals can perform. Quadrupedal animals like leopards can leap and catch birds; dogs can jump in the air and hit the ball using their heads. Usually, this type of acrobatic behavior consists of multiple phases, including jumping, interacting with the target object, and landing phases. Ball bumping motion, which requires the quadruped to jump into the air and hit a falling ball toward a goal location, is one of the most representative acrobatic motions for quadrupeds. Different from standard quadrupedal locomotion on the ground, such a ball bumping motion is highly time-sensitive, i.e., if the juggler does not arrive at the expected position with the specific velocity at a proper time, then the ball can not reach the desired region. In this paper, we aim to tackle the specific ball bumping problem with quadruped robots, which serves as a starting point for exploring more highly dynamic motion planning and control frameworks for future robotic applications.

\subsection{Related Works}
Dynamic locomotion for quadruped robots has attracted a considerable amount of research attention during the past decade. Thanks to the rapid advancements of both hardware and algorithms, quadrupedal robots have demonstrated impressive locomotion skills on flat and uneven terrains~\cite{di2018dynamic, bledt17, bel16, mas17}.

\figdemo

Fundamentally speaking, achieving jumping motion for quadrupeds can be formulated as trajectory optimization problems in general~\cite{kelly17}. MIT Cheetah 3 is capable of jumping onto a desk with a height of 30 inches by an offline trajectory optimization that considers full-body kino-dynamics in a 2D vertical plane \cite{bledt2018cheetah}. Chigonoli \emph{et al.} \cite{chignoli2021rapid} proposed a hierarchical planning framework with centroidal dynamics and joint-level kinematics to achieve Omnidirectional jumping, which takes on average 0.55 s to find a reference trajectory plan. Nguyen \emph{et al.} ~\cite{nguyen2021contact} synthesized a full-body trajectory optimization to achieve 3D jump with quadrupeds, which takes several minutes to solve. Among the pioneering works trying to plan and control jumping motions for quadrupeds, Park \emph{et al.}~\cite{park2015online,park2021jumping} adopted a 2D single rigid body model for quadrupedal dynamics and developed an event-triggered jumping controller that accomplishes jumping over obstacles on MIT Cheetah 2 quadruped. Li \emph{et al.}~\cite{li2021quadruped} studied the jumping motion of quadrupeds with only two rear legs and developed a hierarchical planning and control framework that can be implemented in real-time based on a spring-loaded inverted pendulum model~\cite{chen2019optimal}, which is verified with simulations. As the reversed process of jumping, landing control with quadrupeds has also been considered. Jeon {\em et al.} \cite{jeon22} developed a supervised learning-based warm start interface for nonlinear landing trajectory planning to improve the performance of quadrupedal landing. How to exploit the conservation of angular momentum to help modulate the robot's configuration during the flight phase has also attracted recent research attentions~\cite{rud21,kurtz2022mini}, which further improves the landing performance of quadrupeds. Despite these amazing demonstrations of jumping motion control for quadrupeds, problems studying quadruped jumping with physical interactions with other objects during the flight phase have not been studied adequately in the literature.

More recently, the problem of operating quadrupeds to physically interact with other objects to achieve more complicated tasks has attracted more research attention. Ji \emph{et al.}~\cite{ji22} considered the soccer shooting problem with a quadruped robot and tackled the problem via a hierarchical reinforcement learning strategy. Shi \emph{et al.}~\cite{shi2021circus} adopted a learning-based approach to enable a quadruped robot to manipulate balls with its legs. All these early investigations focus on scenarios where the target object is static or quasi-static. How to design effective planning and control strategies for interaction with moving objects remains lacking. 

In addition to quadrupeds, acrobatic motions for bipedal robots have also been studied extensively in the literature. In particular, investigations on jumping control with bipeds date back to Raibert's seminal works~\cite{raibert1986legged,hodgins1988biped}. Hierarchical planning and control frameworks have been long employed in the synthesis of acrobatic bipedal motions. Wensing \emph{et al.}~\cite{wensing2013generation} proposed a conic optimization-based task-space control to achieve motions like ball kicking and standing broad jump with a humanoid robot. Xiong \emph{et al.}~\cite{xio18} developed an optimization-based controller for biped hopping with biped Cassie. Chigonoli \emph{et al.} \cite{chi21} employed the trajectory optimization-based approach for synthesizing dynamic acrobatic motions with a humanoid robot. Reinforcement learning-based approaches have also demonstrated impressive performance in achieving agile biped behaviors~\cite{siekmann2021sim,yu2022dynamic}. Poggensee \emph{et al.} \cite{poggensee2020ball} implemented ball-juggling on biped Cassie, the motion is a periodic orbit, and there is no flying phase while bounding.

\subsection{Contributions}
The main contributions of this work are summarized as follows. First, to incorporate the challenging time-state constraint in the ball hitting problem, we develop a real-time optimization scheme to find the critical time-state pair based on a two-dimensional single rigid body model. Such an online planning scheme allows us to find the appropriate take-off state and time to achieve precise ball hitting, which addresses the main challenge in accomplishing the ball bumping problem. Second, to enable the quadruped to accurately execute the planned motion, we develop an improved model predictive controller that actively adjusts the ground reaction forces based on the time remaining for the jumping motion. Such a modification effectively guarantees the execution of the time-restricted motion, making the ball bumping motion practically achievable. Third, we demonstrate the effectiveness and performance of the proposed framework on a real quadruped platform, successfully accomplishing the ball bumping motion with various initializations.

\section{Problem Statement}\label{sec:prob}

Our goal is to control the quadruped to bump the falling ball to the desired landing point. Throughout this paper, we assume the initial state of the ball is given, including the released position, velocity, and the released time $t_s$.  As depicted in Fig.~\ref{fig:bbq_process}, the whole process can be split into three phases. The first phase is takeoff, in which the quadruped starts to jump at $t_j$ according to the released time $t_s$ of the ball, then executes the online planned jumping motion and finally takes off at  $t_t$. The second is a flying phase, and the quadruped will fly off the ground based on the takeoff velocity, then collide with the ball at $t_h$ in order to let the ball reach the desired landing region. The last one is the landing phase, where the quadruped will keep balance after contact with the ground.

The ball bumping motion studied in this paper requires an active real-time plan of a feasible trajectory for the quadruped in response to the falling ball's state and accurate execution of the planned trajectory. Such a problem calls for a proper balance between the efficiency and accuracy of the proposed approach. In this section, we provide our modeling of the quadruped-ball system used for planning the ball bumping task and then give an overview of the proposed framework. 
\figprocess

\subsection{Modeling of Jumping Quadruped and the Falling Ball}
\subsubsection{quadruped model}
\begin{figure}[!ht]
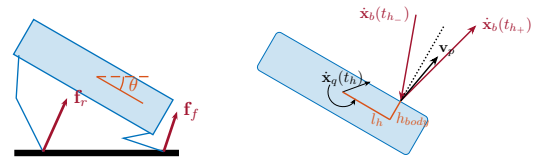

    \centering
    \resizebox{0.45\linewidth}{!}{
        \figplanarmodel
    }
    \resizebox{0.45\linewidth}{!}{
        \fighitvel
    }
    \caption{\textbf{Left:}2D single rigid body model; \textbf{Right:} collision model.}
    \label{fig:planar_model}
\end{figure}

In order to simplify the planning of jumping motion while reserving acceptable accuracy, we consider a 2D planar single rigid body model(2D-SRBM) for quadruped in the sagittal plane. The state is the position of the center of mass(COM) and the pitch angle of the quadruped, denoted as $x_q \in SE(2)$. The dynamics are given as follows
\begin{align}
    \label{eq:planner_dyna}
    \begin{aligned}
    \frac{\textrm{d}}{\textrm{d}t}
    \begin{bmatrix}
        \mathbf{x}_q \\ \dot{\mathbf{x}}_q
    \end{bmatrix}
    &=\begin{bmatrix}
        \mathbf{0}_3 & \mathbf{1}_3 \\
        \mathbf{0}_3 & \mathbf{0}_3 
    \end{bmatrix}
    \begin{bmatrix}
        \mathbf{x}_q \\ \dot{\mathbf{x}}_q
    \end{bmatrix}\\
    &+
    \begin{bmatrix}
        \mathbf{0}_{3 \times 2} & \mathbf{0}_{3 \times 2} \\
        \mathbf{1}_2/m & \mathbf{1}_2/m \\
        I^{-1} [\mathbf{r}_f]_{\times} &  I^{-1} [\mathbf{r}_r]_{\times}
    \end{bmatrix}
    \begin{bmatrix}
        \mathbf{f}_f \\ \mathbf{f}_r
    \end{bmatrix}
    +
    \begin{bmatrix}
        \mathbf{0}_{4\times1} \\ g \\ 0
    \end{bmatrix}
    \end{aligned}
\end{align}
where $[\mathbf{a}]_{\times}$ is defined as an operator such that $[\mathbf{a}]_\times \mathbf{b} = \mathbf{a}_{1}\mathbf{b}_{2} - \mathbf{a}_{2}\mathbf{b}_{1}$ for all $\mathbf{a},\mathbf{b} \in \R^2$, $\mathbf{0}_n$ and $\mathbf{0}_{m\times n}$ are the n by n and m by n zeros matrices respectively. $I \in \R^{2\times2}$ is the inertial matrix in x-z plane. As shown in Fig.~\ref{fig:planar_model}, $\mathbf{r}_f$, $\mathbf{r}_r$ are the vectors from the COM to the front and rear feet respectively. $\mathbf{f}_f$, $\mathbf{f}_r$ are the corresponding ground reaction forces.

\subsubsection{flying ball/quadruped model}\label{sec:flymodel}
Because of the lightweight legs and the short duration of the flying phase, it is hard to adjust the angular momentum of the torso in the air. Therefore, we can consider the position trajectory of the quadruped torso between time $t_t$ and $t_h$ as a ballistic trajectory. As for the rotation of the quadruped, we believe the robot will keep a constant angular velocity after taking off. 

As for the falling ball, if the chosen ball has high-density mass and the velocity is low, then we can neglect the aerodynamic effect. Hence, the motion of the falling ball can also be modeled as a ballistic trajectory before and after the collision.

\begin{equation}
\mathbf{x}_{b/q}(t) 
    =\left\{
    \begin{aligned}
    &\mathbf{x}_{b/q}(0) + \dot{\mathbf{x}}_{b/q}(0)t + \frac{1}{2} \mathbf{g}_{b/q}t^2 &, t < t_h, \\
    &
    \begin{aligned}
        \mathbf{x}_{b/q}(t_h) + \dot{\mathbf{x}}_{b/q}(t_{h_+}) (t-t_h)  \\
        + \frac{1}{2} \mathbf{g}_{b/q}(t-t_h)^2 &
    \end{aligned} &, t > t_h ,
    \end{aligned}
\right.
\label{eq:flight}
\end{equation}
where $\mathbf{x}_b \in \R^2$ is the position of the ball in the x-z plane, $\mathbf{x}_q \in SE(2)$ is the 2D state of quadruped, and $\mathbf{g}_b = [0\ g]^T$, $\mathbf{g}_q = [0\ g\ 0]^T$ are gravity vectors.

\begin{figure*}[tp!]
    \centering
    \includegraphics[width=0.6\linewidth]{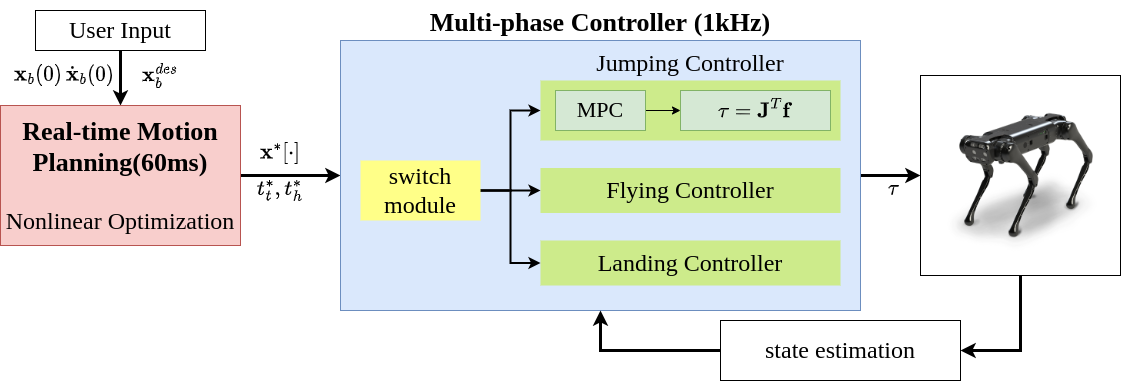}
    \caption{Motion planning and control framework for ball bumping problem}
    \label{fig:framework}
    \vspace{-15px}
\end{figure*}
\subsubsection{quadruped-ball collision model}
The collision model is used to dictate the velocity-changing rules for both the ball and quadruped while the collision occurs. As shown in Fig.~\ref{fig:planar_model}, we assume the collision will not affect the motion along the tangential direction of the contact surface.
According to the conservation of momentum and considering the enormous difference between the mass of the ball and the quadruped, we assume that the quadruped's state will not change after the collision, while the velocity of the ball after the collision is determined as follows
\begin{equation*}
    \dot{\mathbf{x}}_b(t_{h_+}) =  \mathbf{R}_h \left(
    \begin{bmatrix}
        1 & 0\\
        0 & -e
    \end{bmatrix}
    \mathbf{R}_h^T \dot{\mathbf{x}}_b(t_{h_-}) +
    \begin{bmatrix}
        0 & 0\\
        0 & e
    \end{bmatrix}
    \mathbf{R}_h^T \mathbf{v}_p \right),
\end{equation*}
where $t_{h_-}$ and $t_{h_+}$ are the time instants before and after the collision, $\mathbf{R}_h \in \R^{2 \times 2}$ represents the orientation from the body to the world frame at $t_h$, $e$ represents the coefficient of restitution, which is set as $0.8$ in our implementation, $\mathbf{v}_q$ is the contact point's velocity of the robot, which is given by
\begin{equation}
 \mathbf{v}_p =    
 \begin{bmatrix}
    \mathbf{1}_2 & \mathbf{R}_h \begin{bmatrix} h_{body}\\ l_h \end{bmatrix}
\end{bmatrix} \dot{\mathbf{x}}_q(t_h),
\end{equation}
where $\mathbf{1}_n$ is the $n \times n$ identity matrix, $l_h$ is the distance between the collision point and COM in the tangential direction of the torso, and $h_{body}$ is the body height.

\subsection{Overview of the Proposed Framework}
The whole ball bumping motion is achieved through four modules: jumping motion planning, jumping controller, flying controller, and landing controller. Because of the agility of the jumping motion, a nonlinear program is formulated to get a dynamically feasible trajectory. The details of the formulation will be illustrated in \ref{sec:planner}. Then the model predictive control is used for the jumping tracking controller. Finally, when foot contact is detected, the system will use the landing controller to help the quadruped to keep balance.
An overview of this framework is shown in Fig.~\ref{fig:framework}.

\section{Real-time Planning of Ball Bumping}\label{sec:planner}

The key point of this ball bounding problem is whether the quadruped can reach the desired velocity when it takes off according to the ball's desired landing point. Also, the robot needs to react quickly after detecting the ball, so the planning time needs to be as short as possible. A straightforward way of real-time implementation is to find a pre-collision state of the quadruped leading the post-collision ball to fly along its ballistic trajectory into the desired landing point with the minimum takeoff twist criterion. However, if only given the take-off state and time by the ball's landing point, because of the high-resolution requirement of jumping motion, it is hard for a controller to reach the desired state accurately.

Therefore, we also need to consider how the robot can reach the desired takeoff state. Some works used 3D single rigid body model as the dynamics of the quadruped for the motion planning, but considering the real-time requirement of the algorithm, it can not satisfy our requirement. Because of the specificity of our jumping motion, which is bilaterally symmetrical, we choose the 2D single rigid body model instead. To further simplify the problem to accelerate the computation, we also set the jumping start time $t_j$ manually according to the ball released time to guarantee the planning can be computed before the robot begins to move.

Finally, in order to find one possible take-off state configuration and a dynamically feasible trajectory in real time, we formulate the following nonlinear optimization problem
\begin{subequations}
\begin{align}
\min_{\mathbf{t}, \mathbf{p}, \dot{\mathbf{x}}_q(t_h), \mathbf{x}[\cdot], \mathbf{u}[\cdot]}  \quad & \sum_{k=1}^{N-1} \omega_k \mathbf{u}[k]^T\mathbf{u}[k] \quad \textrm{s.t.}\\ 
\textrm{(Dynamics)}             \quad   & \mathbf{x}[k+1] = f(\mathbf{x}[k], \mathbf{u}[k], \delta t) \\
\textrm{(Ball's landing point)} \quad   & \mathbf{x}_b^{des} = p(t_h, \mathbf{x}_b(0), \dot{\mathbf{x}}_b(0), \dot{\mathbf{x}}_q(t_h)) \\
\textrm{(Collision point)}      \quad   & \underbar{$l$} \leq l_h \leq \bar l \label{eq:l}    \\
\textrm{(Takeoff Time)}         \quad   & t_j \leq t_t \leq t_h \label{eq:takeoff_time} \\
\textrm{(Friction cone)}         \quad   & \mathbf{u}[k] \in \mathbf{F}(\mathbf{u}[k]) \label{eq:grf2} \\
\textrm{(Initial conditions)}   \quad   & \mathbf{x}[1] = \mathbf{x}_0  \label{eq:initial} \\
\textrm{(Takeoff conditions)}    \quad  
                                        & \underbar{$\mathbf{x}$} \leq \mathbf{x}[N] \leq \bar{\mathbf{x}}
\end{align}
\end{subequations}
where decision variables $ \mathbf{x}[k]=[\mathbf{x}_q[k], \dot{\mathbf{x}}_q[k]]^T \in \R^6$ and $\mathbf{u}[k] = [f_f^T[k], f_r^T[k]]^T \in \R^6$ is the state and input of the 2D-SRBM respectively, and $ \mathbf{x}[\cdot], \mathbf{u}[\cdot]$ is the state and input trajectories between time $t_j$ and $t_t$. Here $\mathbf{t}=[t_t, t_h]^T$ is the time frames which need to be optimized, $\mathbf{p} = [l_h, \theta_h]^T$ is the collision configuration waited to be optimized. There are totally $ N $ timesteps for this jumping motion, $ \omega_i $ is the cost weight, $f(\mathbf{x}[k], \mathbf{u}[k], \delta t), p(t_h, \mathbf{x}_b(0), \dot{\mathbf{x}}_b(0), \dot{\mathbf{x}}_q(t_h)), \mathbf{F}( \mathbf{u}[k]) $ are the functions for the discrete 2D-SRBM, the ball's landing position, and the friction cone, where $\delta t =\frac{t_t - t_j}{N-1}$ is the timestep for dynamics. And the collision position and takeoff state is bounded by $\underbar{$l$}, \bar l$ and $\underbar{$\mathbf{x}$}, \bar{\mathbf{x}}$ respectively.

\subsection{Ball's Landing Point Design}
From the above formulation, we optimized the collision time $t_h$ and collision state $\dot{\mathbf{x}}_q(t_h)$ at the same time. And they are associated by the desired landing point of the ball.

After the collision, the ball will fall only under the effect of gravity as described at \ref{sec:flymodel}. With the velocity $\dot{\mathbf{x}}_{b}(t_{h_+})$ and position $\mathbf{x}_{b}(t_{h_+})$ of the ball after the collision, we can get the landing point of the ball, which is what we need.

With $z$ axis value of the velocity $\dot{\mathbf{x}}_{b,z}(t_{h_+})$ and position $\mathbf{x}_{b,z}(t_{h_+})$ of the ball after the collision, we can easily get the time $t_e$ when the ball touch the desired ground, the height of the ground is denoted as $\mathbf{x}_{b,z}(t_e)$, then $t_e$ can be computed

\begin{equation}
    t_e = \frac{1}{g} \Biggl( g t_h - \dot{\mathbf{x}}_{b,z}(t_{h_+}) + 
    \sqrt{
    \begin{aligned}
        &(\dot{\mathbf{x}}_{b,z}(t_{h_+}) - g t_h)^2 \\
        &- 2g( \mathbf{x}_{b,z}(t_h) - \dot{\mathbf{x}}_{b,z}(t_{h_+}) t_h \\
        &+ \frac{1}{2} g t_h^2 - \mathbf{x}_{b,z}(t_e))
    \end{aligned}
    } \Biggl) ,
\end{equation}

Then using the dynamics of the ball (\cref{eq:flight}), we can get the position function for the ball's landing point $\mathbf{x}_b(t_e)$, which depends on $t_h$, $\mathbf{x}_b(0)$, $\dot{\mathbf{x}}_b(0)$ and $\dot{\mathbf{x}}_q(t_h)$,
\begin{equation}
    \label{eq:landing_point}
    \mathbf{x}_b(t_e) = p(t_h, \mathbf{x}_b(0), \dot{\mathbf{x}}_b(0), \dot{\mathbf{x}}_q(t_h)).
\end{equation}

Then because during the planning, we only consider the trajectory from time $t_j$ to $t_t$, so here we need to induce $\mathbf{x}_q(t_h)$ by $\mathbf{x}_q(t_t)$ using the model described at \ref{sec:flymodel}
\begin{align}
    \mathbf{x}_q(t_h) &= 
    \begin{bmatrix}
        \mathbf{x}_b(t_t) - \mathbf{R}_h \begin{bmatrix} l_h \\ r_h \end{bmatrix}   \\
        \theta_h
    \end{bmatrix}
    - \dot{\mathbf{x}}_q(t_t) \Delta t
    - 0.5 \mathbf{g}_q \Delta t^2    \\
    \dot{\mathbf{x}}_q(t_h) &= \dot{\mathbf{x}}_q(t_t) - \mathbf{g}_q \Delta t
\end{align}
where $\Delta t = t_h - t_t$, $\mathbf{g}_q = [0\ g\ 0]^T$ is the gravity.

Here we formulate the ball's landing point task as a constraint not cost, the main reason is that the jumping task is time-sensitive, so it's better to add this task into the constraint to guarantee the reference trajectory can lead the quadruped to reach the desired takeoff state accurately at the desired time instance. Besides, as for the computational cost and success rate, we verified that the success rate of putting the task into the constraint or the cost was roughly the same, and they all spent nearly 60 ms to 80ms.

\subsection{Optimization Objective Design}
Because we only need to get a feasible solution, so we only choose the input norm $\mathbf{u}[k]^T\mathbf{u}[k]$ as the optimization cost, which means if the problem has a solution, whatever it is globally optimal, it can be used for our jumping motion.

\section{Multi-phase Controller Design for Ball Bumping}\label{sec:controller}

As shown in Fig.~\ref{fig:framework}, there are three main phases in this ball bumping task, in order to achieve the desired landing point accurately and keep the whole process safe, we customize the controllers for all these three phases respectively.

\subsection{Jumping Controller}\label{sec:jump_ctrl}

The jumping motion is the most important part of the ball bumping task, and its focus is different from other classical locomotion tasks of quadrupeds, like walking on narrow terrains. Because jumping motion changes the velocity in small duration, which means it is a high-resolution motion for the controller to track. Besides, the most critical challenge is that jumping motion in the ball bumping task is highly time-sensitive. The quadruped needs to reach the desired takeoff state accurately at a specific time; otherwise, the ball will deviate a lot from the goal. According to this property, there should be a strict time-state constraint imposed on the take-off state.

To fulfill the requirements of the jumping motion, we modify the famous MPC controller\cite{di2018dynamic} for our problem.

\begin{subequations}
\begin{align}
\min_{\mathbf{x}_c, \mathbf{u}_c}  \quad & \sum_{i=1}^{K-1} \lVert \mathbf{u}_c[i] \rVert_{\mathbf{W}} + \lVert \mathbf{x}_c[i]- \mathbf{x}_{ref}[i]) \rVert_{\mathbf{Q}} \label{eq:cost} \\
                    & + \lVert \mathbf{x}_c[K]- \mathbf{x}_{ref}[K] \rVert_{\mathbf{Q}_p} \nonumber \\ 
\textrm{s.t.} \quad & \mathbf{x}_c[i+1] = \mathbf{A}\mathbf{x}_c[i]+ \mathbf{B} \mathbf{u}_c[i] \label{eq:dynamics} \\
                    & \underbar{$\mathbf{c}$} \leq \mathbf{C} \mathbf{u}_c[i] \leq \bar{\mathbf{c}} \label{eq:grf3}
\end{align}
\end{subequations}
where $\mathbf{x}_c[i] = [\mathbf{\Theta}^{T}\ \mathbf{p}^{T}\ \boldsymbol{\omega}^{T}\  \dot{\mathbf{p}}^{T}\ g]^{T}$, four ground reaction forces produced by legs is the system input $\mathbf{u}_c[i] = [\mathbf{f}_1^T \cdots \mathbf{f}_4^T]^T$,  $\boldsymbol{\Theta}= [\phi\ \theta\ \psi]^T$ denotes the Euler angles representing the quadruped's orientation. $\mathbf{A} \in \R^{13 \times 13}$ and $\mathbf{B} \in \R^{13 \times 12}$ are the parameter matrices of discrete 3D-SRBM. 
Compared to the standard formula of MPC, we transfer the strict time-state constraints of the jumping motion into the cost, which is set as a terminal cost for the takeoff state.

\subsubsection{Effects of the Terminal Cost}
The performance of the convex MPC has been validated in the MIT Cheetah 3 \cite{di2018dynamic} for high-speed running. However, for the ball bumping motion, which is time-sensitive, we need to make sure the quadruped body can reach the desired position and velocity at the desired time instance.

To solve this issue, we add a terminal cost $ \lVert \mathbf{x}_c[k]- \mathbf{x}_{ref}[k] \rVert_{\mathbf{Q}_p} $ to help the robot to reach the desired hitting state in time. Both $ \mathbf{Q_p} $ and $ \mathbf{Q} $ are diagonal matrices, every diagonal entries of $ \mathbf{Q_p} $ need to greater than $ \mathbf{Q} $'s relatively. This terminal cost can help the controller finds a sequence of control inputs that will guide the system to reach the terminal state at the exact time $t_t$ rather than ahead of $t_t$.

\subsubsection{Dynamic Horizon Tuning}
During the trajectory tracking process, we will tune the horizon and timestep $ \textrm{d}t $ of the discrete SRBM in order to make sure $\mathbf{x}_q^*(t_t)$ computed by the previous motion planner is always the terminal reference state for the jumping controller.
We also use horizon cutting and dynamic $\textrm{d}t$ to make sure $\mathbf{x}_q^*(t_t)$ is always the terminal reference state of the controller. This tuning method is described in Algorithm \ref{alg: hori}.

\begin{algorithm}
\caption{Dynamic horizon tuning}
\label{alg: hori}
\KwData{current time $t_{now}$, takeoff time $t_t$, horizon $h$, minimum time interval $\textrm{d}t_{min}$ }
\KwResult{$\textrm{d}t$, $h$}
$\textrm{d}_t \leftarrow (t_{now} - t_t)/h $ \;
\If{$\textrm{d}t < \textrm{d}t_{min}$}{
    $\textrm{d}t \leftarrow \textrm{d}t_{min} $ \;
}
\For{$i\leftarrow 1$ \KwTo $h$}{
\If{$t_{now} +i \textrm{d}t > t_t$}{
    $h \leftarrow i $ \;
}
}
\end{algorithm}

Finally, the revised MPC problem can be formulated as a quadratic programming problem and quickly solved. Then with the foot jacobian of each leg $\mathbf{J}_i \in \R^{3\times3} $, joint torques of each leg $i$ can be given by 
\begin{equation}
   \boldsymbol{\tau}_i = \mathbf{J}_i^T \mathbf{R}^T \mathbf{f}_i.
\end{equation}

\subsection{Flying Controller}
As described in \ref{sec:flymodel}, the motion trajectory between the takeoff and collision is considered a parabola. Then, to reduce the effect of the leg motion further, we freeze all legs during this process, which means we use the PD control to keep all joint angles constant.

After the collision $t_h$ and a small duration of delay, in order to adjust the leg configuration for landing, we switch to the falling controller; the controller uses Cartesian impedance control:
\begin{equation}
    \boldsymbol{\tau}_i = \mathbf{J}_i^T[\mathbf{K}_p(\mathbf{p}_i^{des} - \mathbf{p}_i) + \mathbf{K}_d(\mathbf{v}_i^{des} - \mathbf{v}_i)]
\end{equation}
to move the foot to desired locations given by
\begin{equation}
    \mathbf{p}^{des}_i = \mathbf{p} + \mathbf{p}^0_i,
\end{equation}
where $\mathbf{p}$ is the robot COM position in the world frame and $\mathbf{p}^0_i$ is the foot position relative to the COM, its $x,y$ axis values are set manually for a stable landing configuration, and the $z$ axis value is the same as the relative $z$ position at the moment of takeoff.

\subsection{Landing Controller}
When any leg contact with the ground is detected by the measurement of the foot contact sensor, the controller uses joint space PD control to damp the robot and finally uses MPC described in \cite{di2018dynamic} to squat and lie on the ground.

\section{Experimental Results and Discussions} \label{sec:exper}

\subsection{Robot Platform}
The whole motion planning and control framework was validated on the real robot platform Aliengo\cite{Aliengo}. Aliengo from Unitree Robotics is a quadruped robot with 12 actuators, whose legs are designed as open chains, and it has a mass of about 22 kg with lightweight legs, so this mechanical design is suitable for the single rigid body model (SRBM) in our methodology. The physical parameters of Aliengo can be found at \cite{Aliengo}.

\subsection{Experiment Setup}
\begin{figure}[b]
    \centering
    \includegraphics[width=0.6\linewidth]{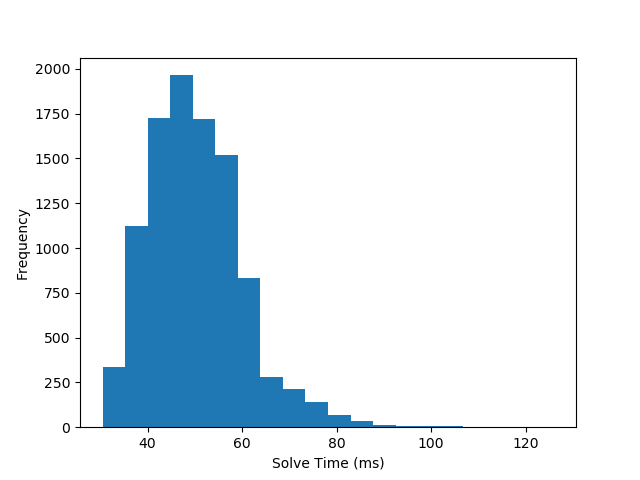}
    \caption{Solving time of 9984 successful solutions, all cold-start.}
    \label{fig:solve_time}
\end{figure}
\begin{figure*}[t!]
    \centering
    \includegraphics[width=\linewidth]{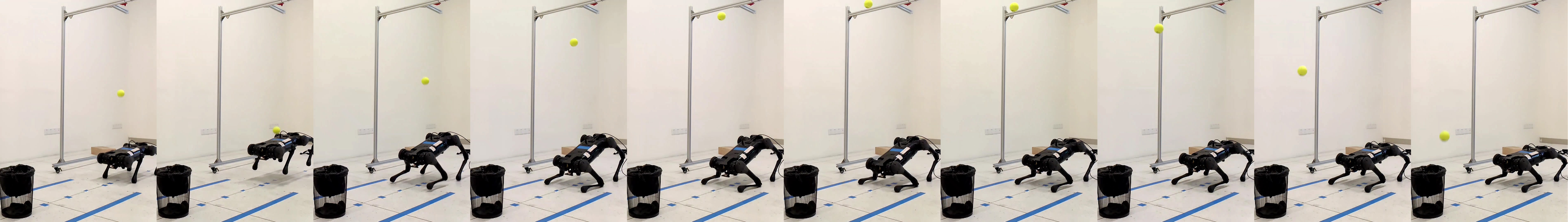}
    \caption{Time series of the ball bumping experiment }
    \label{fig:exp_bumping}
\end{figure*}
\plottraj
\figforce
The motion planning and control framework was developed with \verb|C++| based on \verb|ROS| and \verb|ros_control| \cite{ros_control}, which provide hardware interface and controller manager and make simulation (using Gazebo \cite{koenig2004design}) and debugging efficiency. \verb|pinocchio|\cite{pinocchioweb} and \verb|Eigen| are used as basic kino-dynamics interfaces in the implementation of the methodology. As for the nonlinear optimization programming for the planner, \verb|CasADi|\cite{andersson2019casadi} was used for constructing the trajectory optimization problem, and \verb|IPOPT| solver was used for solving it. Quadruped's body mass $m$ was set to 35 kg for increased gains of the system due to the large latency (about 8 ms loop-back) caused by \verb|unitree_legged_sdk|.

The hardware setup is shown in Fig.~\ref{fig:exp_demo_setup}; the software described above was running on a laptop with an i7-8550U processor and runs Ubuntu Linux with a kernel patched with \verb|CONFIG_PREEMPT_RT|. Aliengo was controlled by UDP through an Ethernet cable with a 1000 Hz control loop. We made a tennis ball releaser to guarantee the ball's released time, position, and velocity. The ball releaser was controlled by the laptop with a USB to CAN cable.


\subsection{Experimental Results}
\subsubsection{NLP Test}
In order to test the performance of the NLP solver for the planner, we did 10000 test cases with 10 evenly distributed samples per interval over the initial state and desired state $\mathbf{x}_{b,1}(0) \in [0, 0.5]$, $\mathbf{x}_{b,2}(0) \in [1.5, 2.0]$,  $\mathbf{x}_{b,1}^{des} \in [0.5, 1.5]$, $\mathbf{x}_{b,2}^{des} \in [0, 0.5]$. All NLPs were solved with a cold start, and the initial guess is set as $ \mathbf{t}^0 = [0.2, 0.4], \mathbf{p}^0 = \mathbf{0}_{2\times 1}, \dot{\mathbf{x}}_q(t_h)^0 = \mathbf{0}_{3\times 1}, \mathbf{x}^0[i] = \mathbf{0}_{3\times 1}, \mathbf{u}^0[i] = [ 0, 0, 175, 0, 0, 175]^T$. The distribution of the solve time is shown in Fig.~\ref{fig:solve_time}, the solve time of failed solutions is excluded, and the success rate is about 99.84\%. The average solving time is about 50ms, most of the solve time is around 80ms, and the maximum solve time is under 120ms, allowing for real-time solving after the ball is released.


\subsubsection{Ball bumping Test}
\begin{table}[thpb]
\centering
\begin{tabular}{|l|ccc|cc|cc|}
\hline
Released point $x_b^0$ /m   & \multicolumn{3}{c|}{0.}                                    & \multicolumn{2}{c|}{0.25}      & \multicolumn{2}{c|}{0.4}       \\ \hline
Desired point $x_{b}^{des}$ /m & \multicolumn{1}{c|}{0.5} & \multicolumn{1}{c|}{0.75} & 1.0 & \multicolumn{1}{c|}{1.0} & 1.5 & \multicolumn{1}{c|}{1.5} & 2.0 \\ \hline
\end{tabular}
\caption{Ball bumping experiment setting}
\label{tab:exp}
\end{table}

We tested 7 cases (see \cref{tab:exp}) with the same initial ball release height $z =  1.56 m$ and different horizontal positions $x$. The whole experiment results are included in the supplemental video. The actual ball landing point in all tests is within a circle with a radius of 10 cm centered on the desired landing point. 

Fig.~\ref{fig:exp_bumping} shows the snapshot of the case with initial ball position $x=0$ m,$ z=1.56$ m and desired landing point $x = 2.5$ m, $z = 0$ m, Fig.~\ref{fig:exp_traj} shows the corresponding pose and twist trajectories given by state estimation module. The ball was released at 0.262 s, and the time instances of a jump start and takeoff were given by solving the nonlinear optimization problem described in \cref{sec:planner}. The NLP solver spent about 60ms to get the optimized time instances. The quadruped started jumping at 0.342 s and took off at 0.691 s. Finally, the robot reached the desired takeoff state at 0.691 s with the jumping controller described in \cref{sec:jump_ctrl}. Fig.~\ref{fig:exp_force} shows the foot force computed by MPC in the jumping process. Although the trajectory has not always been tracked very well, the error of takeoff state is acceptable relatively. The state estimation data can not be trusted after the jumping because the foot is off the ground.

\section{Conclusions and Future Work}\label{sec:conclude}
This paper presented a motion planning and control framework for the ball bumping task that leads the quadruped bump of the ball into the desired landing region. A nonlinear optimization problem is formulated for a dynamically feasible jumping trajectory, which can be solved online. The multi-phase controller is designed for this time-sensitive trajectory, which can reach the desired takeoff state at desired time instance accurately. And the performance has been validated on the Aliengo.

Future work can be devoted to the warm-start interface of the nonlinear optimization problem to get a more reasonable initial guess. A vision detection and tracking system can be developed and run onboard, allowing the dropping of the ball by human hand without the releaser. In addition, the planning for landing motion can be developed for a more steady landing performance.

\newpage
\bibliographystyle{ieeetr}
\bibliography{mybibfile}

\end{document}